\documentclass[11pt]{article}
\usepackage{times}
\usepackage{fullpage}
\usepackage{latexsym}
\usepackage{epsf}
\usepackage{latexsym}
\usepackage{amsmath}

\newcommand{\bi}{\begin{itemize}}
\newcommand{\ei}{\end{itemize}}
\newcommand{\be}{\begin{enumerate}}
\newcommand{\ee}{\end{enumerate}}
\newcommand{\bd}{\begin{description}}
\newcommand{\ed}{\end{description}}
\newcommand{\bt}{\begin{tabbing}}
\newcommand{\et}{\end{tabbing}}
\newcommand{\beq}{\begin{equation}}
\newcommand{\eeq}{\end{equation}}
\newcommand{\vsone}{{\em 1-vs-1}}
\newcommand{\vsother}{{\em 1-vs-other}}

\newcommand{\sect}[1]{Section~\ref{#1}}
\newcommand{\tbl}[1]{Table~\ref{tab:#1}}

\newcommand{\equat}[1]{Eq.(\ref{eq:#1})}

\newcommand{\slmax}{SL1-Max}
\newcommand{\boost}{Adaboost}

\newcommand{\x}{\mathbf{x}}
\newcommand{\fvec}{\mathbf{f}}
\newcommand{\lbd}{\boldsymbol{\lambda}}
\newcommand{\cl}{c}
\newcommand{\clp}{d}
\newcommand{\sumi}{\sum\limits_i}
\newcommand{\sumdj}{\sum\limits_{d,j}}
\newcommand{\sumj}{\sum\limits_{j}}
\newcommand{\sumcl}{\sum\limits_{\cl}}
\newcommand{\sumx}{\sum\limits_{\x}}
\newcommand{\sumxcl}{\sum\limits_{\x,\cl}}
\newcommand{\sumicl}{\sum\limits_{i,\cl}}
\newcommand{\feat}{f_{\clp,j}}
\newcommand{\betaj}{\beta_{\clp,j}}
\newcommand{\Pest}{p}
\newcommand{\Pgibbs}{q_{\lbd}}
\newcommand{\Ogibbs}[1]{e^{\lbd \cdot \fvec(#1)}}
\newcommand{\Z}{Z_{\lbd}}
\newcommand{\Ptrue}{p}
\newcommand{\Pemp}{\tilde{p}}
\newcommand{\Semp}{\tilde{\sigma}}
\newcommand{\xset}{X}

\newcommand{\cvec}{\mathbf{y}}
\newcommand{\celem}{y}
\newcommand{\Nx}{m}
\newcommand{\Nfeat}{n}
\newcommand{\Nclass}{l}
\newcommand{\argmax}{\mathop{\rm argmax}}
\newcommand{\entr}[1]{H(#1)}
\newcommand{\myand}{\mathrm{\ and\ }}
\newcommand{\abso}[1]{\left|#1\right|}
\newcommand{\maxl}{\max\limits}
\newcommand{\minl}{\min\limits}
\newcommand{\loss}{L^{\boldsymbol{\beta}}_{\Pemp}}

\newcommand{\optp}[3]{
$$
\begin{array}{rlrl}
#1: & #2 \mathrm{\ subject\ to}\\
& \left\{ \begin{array}{l} #3 \end{array}\right.
\end{array}
$$
}

\newcommand{\optq}[3]{
$$
\begin{array}{rlrl}
#1: & #2 \mathrm{\ with}\\
& \left\{ \begin{array}{l} #3 \end{array}\right.
\end{array}
$$
}

\newcommand{\pjoint}{\mathcal{P}_1}
\newcommand{\pclass}{\mathcal{P}_2}
\newcommand{\ppost}{\mathcal{P}_3}
\newcommand{\qjoint}{\mathcal{Q}_1(\lbd)}
\newcommand{\qjointp}{\mathcal{Q}_1(\lbd^{\cl})}
\newcommand{\qclass}{\mathcal{Q}_2(\lbd_{\cl})}
\newcommand{\qpost}{\mathcal{Q}_3(\lbd)}
\newcommand{\qpostp}{\mathcal{Q}_3(\lbd^{\cl})}

\newcommand{\ignore}[1]{}

\newcommand{\emcite}[1]{\cite{#1}}

\title{Efficient Multiclass Implementations of L1-Regularized Maximum Entropy}
\author{
\begin{tabular}[t]{cc}
Patrick Haffner \quad Steven Phillips & Rob Schapire \\
AT\&T Labs -- Research & Princeton University\\
180 Park Avenue, Florham Park, NJ 07932 & 35 Olden Street, Princeton, NJ  08544\\
\it \{haffner, phillips\}@research.att.com & \it schapire@cs.princeton.edu\\
\end{tabular}
}

\date{}

\begin{document}
\maketitle

\begin{abstract}
This paper discusses the application of L1-regularized maximum entropy
modeling or {\em SL1-Max}~\cite{maxent-colt} to multiclass categorization
problems. A new modification to the SL1-Max fast sequential learning algorithm
is proposed to handle {\em conditional} distributions. Furthermore, unlike most
previous studies, the present research goes beyond a single type of
conditional distribution. It describes and compares a variety of modeling
assumptions about the class distribution (independent or exclusive) and
various types of joint or conditional distributions.  It results in a new
methodology for combining binary regularized classifiers to achieve multiclass
categorization. In this context, Maximum Entropy can be considered as a
generic and efficient regularized classification tool that matches or
outperforms the state-of-the art represented by AdaBoost and SVMs.

\end{abstract}
\section{Introduction}

A new form of maximum entropy (maxent) with a {\em sequential} updating
procedure and {\em L1}
regularization (\slmax) was recently introduced~\cite{maxent-colt} as a
probability distribution estimation technique. This study adapts \slmax\ to
classification problems.  It demonstrates  a regularized linear classification
algorithm that bears striking similarities with large margin classifiers such
as AdaBoost~\cite{schapire99,collins00}.  

Conditional maxent models~\cite{berger} (also known as conditional exponential
or logistic regression models) were previously applied to classification
problems in text classification~\cite{nigam}. These models were
shown~\cite{jaakkola99} to be a generalization of Support Vector Machines
(SVMs)~\cite{vapnik98} or a modification of AdaBoost normalized to form a
conditional distribution~\cite{lebanon}. The three aforementioned references
employed the L2 type of regularization. L1 regularization was proposed for
logistic regression~\cite{ng}, which is a particular case of maximum entropy.
The application of conditional maxent to part-of-speech tagging or machine
translation problems~\cite{och} can also be seen as a classification problem,
where the number of classes is very large. Solutions dealing with the
computational and memory usage issues arising from this large number of
classes were proposed for translation applications.  Most of these studies
focus on {\em specific} applications.  We could not find studies of maxent as
a generic classification algorithm that can be applied to a wide range of
problems.

This situation can be contrasted to the literature on large margin
classifiers, where extant studies~\cite{allwein} cover their adaptation to
multiclass problems. Large margin classifiers were initially demonstrated on
binary classification problems and extended to multiclass classification
through various schemes combining these binary classifiers. The simplest
scheme is to train binary classifiers to distinguish the examples belonging to
one class from the examples not belonging to this class.  This approach is
usually referred to in the literature as \vsother\ or {\em 1-vs-all}.  Many
other combination schemes are possible, in particular \vsone\ where each
classifier is trained to separate one class from another. More general
combination schemes include error correcting output codes
(ECOC)~\cite{crammer00} and hierarchies of classifiers.  \ignore{A unified
framework~\cite{allwein} for studying these different techniques to reduce a
multiclass categorization problem to binary classifiers provides
generalization bounds.  With methods based on recombination of binary
classifiers, each classifier corresponds to a different optimization problem
and is not guaranteed to be globally optimal. ~\cite{schapire99} describe and
compare various global multiclass implementations of AdaBoost.}

Our goal is to include maxent among the regularized classification algorithms
one would routinely consider, and implement it in a software package that
would be as easy to use as SVMs and Adaboost packages. The expected advantage
of maxent over other classification algorithms is its flexibility, both in
terms of choice of distribution and modeling assumptions. \slmax\ provides the
ideal starting point for this work: this algorithm estimates maxent model
parameters in a fast sequential manner, and supports an effective and well
understood L1 regularization scheme (which leads to sparser solutions than L2
regularization).  The new contributions in this paper are the following. We
adapt \slmax\ to conditional distributions, which requires the derivation on a
new bound on the decrease in the loss. We compare the joint and the
class-conditional distributions to the conditional distribution traditionally
considered in the literature. We introduce ``non-class'' maxent models to
reduce multiclass problems to a set of binary problems (to our knowledge, such
techniques have only been used in question answering
systems~\cite{nonclass}).  We show, through experiments, that maxent
statistical interpretation leads to a new methodology for selecting the
optimal multiclass approach for a given application.

\sect{def} introduces the notation to handle classification problems.
In \sect{general}, we adapt the \slmax\
algorithm to estimate parameters of the joint, class-conditional and
conditional distributions. In \sect{multi}, we generalize these techniques to the multi-label case.
After a discussion about the implementation in \sect{implement}, comparative experiments are provided in \sect{expe}.

\ignore{
focus on \vsother so far
feature selection

\slmax\ has only been introduced recently~\cite{maxent-colt} and no study on how
to apply it efficiently to multiclass classification is yet available. There
are several ways to
incorporate the concept of class in a data-generating approach such as
\slmax, in particular by estimating either the conditional~\cite{nigam} or the
class conditional distributions.

Its application to classification
problems allows us to explicitly model observations that other classification
approaches would leave as open problems.
\bi
\item How to reject test data that belongs to none of the classes that
  characterized the training data.
\item How to model the fact that classes are exclusive or that they can
  overlap (when an example can belong to several classes).
\item Understanding when a combination of  binary classifier should work as
  well as global system.
\ei
While maxent is an old technique, is seems to provide new insight on
how to optimize an algorithm on multiple classes.
}

\section{Definitions and notation}
\label{def}
Our sample space covers input-label associative pairs $(\x, \cl) \in \xset \times
\{1, \dots, \Nclass\} $. Our goal is to determine, for a given input $\x$, the most likely
class label $\cl^*$ which maximizes the unknown conditional distribution
$\cl^* = \argmax_\cl \Ptrue(\cl|\x)$. For simplicity, this paper initially
focuses on classification where each input is associated with a single
label. From a statistical viewpoint, this means that classes are {\em
  exclusive} (i.e. they cannot occur simultaneously).
Models where multiple labels are allowed will be considered later.

The application of Bayes Rules writes
\beq
\Ptrue(\cl|\x)= \frac{\Ptrue(\x,\cl)}{\Ptrue(\x)} = \Ptrue(\cl)
\frac{\Ptrue(\x|\cl)}{\Ptrue(\x)}.
\eeq
As $\Ptrue(\x)$ does not impact
the choice of the class, we can choose which distribution we want to estimate:
{\em joint} with $\Ptrue(\x,\cl)$, {\em
  conditional}  with $\Ptrue(\cl|\x)$, or
{\em class-conditional} with $\Ptrue(\x|\cl)$. The rest of this section 
introduces notation to manipulate these distributions in a
consistent fashion.

In maxent, training data is used to impose constraints on the
distribution. Each constraint expresses a characteristic of the training data
that must be learned in the estimated distribution $\Pest$. Typical constraints require
features to have the same expected
value as in the training data.
Features are real valued functions of the input and of the class  $f(\x,
\cl)$. To represent the average of a  feature $f$ over a distribution $\Pest$, we use the following notation:
\bd
\item[Joint:] The expected value of $f$ under $\Pest$ is $$\Pest[f]= \sumxcl
  \Pest(\x,\cl) f(\x, \cl).$$
\item[Conditional:] For a given training example $\x_i$,  $\Pest_i(\cl)=\Pest(\cl|\x_i)$ and
   $$\Pest_i[f]= \sumcl \Pest(\cl|\x_i) f(\x_i, \cl)= \sumcl \Pest_i(\cl) f(\x_i, \cl).$$
\item[Class-conditional:] For a given class $\cl$,  $\Pest_{\cl}(\x)=\Pest(\x|\cl)$ and
   $$\Pest_{\cl}[f]= \sumx \Pest(\x|\cl) f(\x, \cl)= \sumx \Pest_\cl(\x) f(\x, \cl).$$ 
\ed

The training data is a set of input-label pairs $(\x_1, \cl_1), \dots,
(\x_{\Nx}, \cl_{\Nx})$, and is a subset of the sample space 
$\{\x_1, \dots, \x_\Nx\} \times \{1, \dots, \Nclass\}$.
The empirical distributions over this training set are defined as follows:
\begin{align*}
\Pemp(\x,\cl)&  =\frac{1}{\Nx}\left|\{1\leq i \leq \Nx: \x_i = \x \myand \cl_i=\cl\} \right|\\
\Pemp_\cl(\x) & = \frac{\left|\{1\leq i \leq \Nx: \x_i = \x \myand \cl_i=\cl\} \right|}{\left|\{1\leq i \leq \Nx: \cl_i=\cl\}
    \right|}\\
\Pemp(\cl)&  = \frac{1}{\Nx}\left|\{1\leq i \leq \Nx: \cl_i=\cl\}\right|
\end{align*}

All maxent models are based on the computation of a linear score over the
features, represented by the inner product between the feature vector and the
weight vector $\lbd$. In the classification case, the feature vector
$\fvec(\x, \cl)$ is defined over all input-label pairs $(\x, \cl)$. This pair
is scored with the inner product $\lbd \cdot \fvec(\x, \cl)$ and compared to
other pairs $(\x, d)$ with $d \neq \cl$.
A subtlety that arises in the application of maxent to classification problems
is the need to multiply each feature as many times as there are classes.
Suppose the input $\x$ is a list of $n$ values $v_1(\x), \dots,
v_n(\x)$, the class dependent features  are defined as follows:
\beq
\label{eq:simpli}
\feat(\x,\cl)= \left\{
\begin{array}{ll}
v_j(\x) & {\rm if\ } \cl= \clp \\
0 & { \rm otherwise}
\end{array}
\right.
\eeq
With this representation,the inner product between the feature vector $\fvec(\x, \cl)$
and the parameter vector $\lbd$ simplifies as
$$
 \lbd \cdot \fvec(\x, \cl)=\sumdj \lambda_{\clp,j} \feat(\x, \cl)=\sumj \lambda_{\cl,j} v_{j}(\x) = \lbd_\cl \cdot
\mathbf{v}(\x),
$$
where  $\lbd_\cl$ is the subset of parameters specific to class $\cl$.

\section{Trying Different Distributions}
\label{general}

This section provides \slmax\ solutions for the estimation of the {\em joint},
{\em class-conditional} and {\em conditional} distributions. It also shows some
limitations of these solutions that will be overcome in the next section. The
first subsection focuses on joint distribution $\Pest(\x,\cl)$.

\subsection{Estimating joint distributions}

\ignore{In this section, we show how to use the \slmax\ algorithm to estimate the parameters of the
maxent model for the joint distribution $\Pest(\x,\cl)$.}

Maximum  Entropy restricts the trained model distribution $\Pest$ so that each
feature has the same expected and empirical means.
\ignore{\beq
\frac{1}{\Nx} \sumi \feat(\x_i, \cl_i) = \sumx \sumcl \Pest(\x,\cl)\feat(\x, \cl).
\eeq}
Our notation summarizes this constraint as $\Pest[\feat] =
\Pemp[\feat]$. In the regularized case, this constraint is softened to have the
form $\abso{\Pest[\feat]-\Pemp[\feat]} \leq \betaj$, where $\betaj$ is a
regularization parameter.

Within these constraints, we are looking for the distribution which is the
closest to the uniform distribution, by maximizing the entropy 
$\entr{\Pest} = -\sumicl \Pest(\x_i,\cl) \ln \Pest(\x_i,\cl)$. This corresponds to
the convex program:
\optp{\pjoint}
{ \maxl_{\Pest}\entr{\Pest}}
{
\sumicl \Pest(\x_i, \cl_i)=1\\
\forall \clp,j: \abso{\Pest[\feat] - \Pemp[\feat]} \leq \betaj
}
The dual program maximizes the likelihood over the exponential distributions.
\optq{\qjoint}
{\minl_{\lbd} \loss(\lbd)} 
{
\loss(\lbd)= 
-\underbrace{\Pemp[\ln{\Pgibbs}]}_{Likelihood}
+\underbrace{\sumdj \betaj \abso{\lambda_{\clp,j}}}_{Regularization}\\
\Pgibbs(\x,\cl)= \frac{1}{\Z}\Ogibbs{\x, \cl}\\
\Z=\sumcl \sumi \Ogibbs{\x_i, \cl}
}

Note that for the joint distribution, the $\Z$ normalization is performed over all classes and all
training samples.
\emcite{maxent-colt} prove the convergence of a sequential-update
algorithm that modifies one weight at a time. This coordinate-wise descent is
particularly efficient when dealing with a large number of sparse features.
A bound on the decrease in the loss is
\begin{multline}
\label{eq:bound}
\loss(\lbd')-\loss(\lbd) \leq -\delta \Pemp[\feat] + \ln\left(1+(e^\delta-1)
\Pgibbs[\feat]\right)\\
 + \betaj \left(\abso{\lambda_{\clp,j}+\delta} - \abso{\lambda_{\clp,j}}\right),
\end{multline}
with equality if we have binary features.
The values of $\delta$ that minimize this expression can be obtained in a
closed form. Note that this analysis must be repeated for all features $j$ and all classes $\clp$.

Efficient implementations of the sequential-update algorithm require storing
numerous variables and intermediate computations. For instance, we
need to store all the $\Pgibbs(\x,\cl)$ and $\lbd \cdot \fvec(\x_i,
\cl_i)$. The storage requirement in $O(m \times n)$ can be problematic for
large-scale problems.

As a matter of fact, we found memory requirements to be the main limitation
of this implementation of multiclass \slmax. Speedup techniques based on partial
pricing strategies~\cite{pricing} have reduced the learning time of \slmax\
and made it manageable.

\ignore{One of these speedups relies on the observation of \equat{simpli}, which
means that $\Ogibbs{\x, \cl}$ does not depend on $\lambda_{\clp,j}$ when $\clp
\neq \cl$. When updating $\lambda_{\clp,j}$ for all others $\cl
\neq \clp$, the $\Ogibbs{\x_i, \cl}$ do not require to be recomputed.}

\subsection{Estimating  class-conditional distributions}
\label{classcond}

The motivation for using the class-conditional distribution $\Pest(\x|\cl)$
is that it allows to build one model per class.
\ignore{If we take into account \equat{simpli},
the averaging of the features based on class conditional distributions
implies the following:
$$
\forall \clp \neq \cl : \Pest_\cl[\feat]=0
$$}
From \equat{simpli}, it is easy to see that for $\clp \neq \cl$, features $\feat$ have no impact on
the class-conditional distribution $\Pest_c$ we are trying to
estimate. As a result, separate optimization problems can be defined for the
$\Nclass$ classifiers with no interaction between
them. 
For each class, the convex  problem $\pclass$ and its convex dual  $\qclass$  are:
\optp{\pclass}
{ \maxl_{\Pest_{\cl}}\entr{\Pest_{\cl}}}
{
\sumi \Pest_{\cl}(\x_i)=1\\
\forall j: \abso{\Pest_{\cl}[f_{\cl,j}] - \Pemp_{\cl}[f_{\cl,j}]} \leq
\beta_{\cl,j}
}
\optq{\qclass}
{\minl_{\lbd_{\cl}} \loss(\lbd_{\cl})}
{
\loss(\lbd_{\cl}) = -\Pemp_{\cl}[\ln{\Pgibbs}]+\sumj \beta_{\cl,j} \abso{\lambda_{\cl,j}}\\
\Pgibbs(\x, \cl)= \frac{1}{\Z(\cl)}\Ogibbs{\x,\cl}\\
\Z(\cl)=\sumi \Ogibbs{\x_i, \cl}
}

For each class $\cl$, we have an independent optimization problem to solve. For each of these
optimization problems, we have a general
solution which is just \slmax.

A clear advantage of the {\em class-conditional} over the {\em joint}
distribution approach is that, when optimizing $\Pest_{\cl}$, we do not have to store the variables used to
optimize the class-conditional distributions of the other classes. This
provides huge savings in memory (i.e the memory requirement is divided by the number of classes).

A  drawback of the class-conditional approach is that it does not minimize
explicitly the classification error rate. To obtain the recognized class, it
relies on the application of the Bayes rules, and thus on the fact that the
probability distributions have been properly estimated. 
Taking the logarithm of $\argmax_\cl \Pemp(\cl)\Pgibbs(\x, \cl)$ , this class is
\footnote{Note that this relies on a good estimation of the $\Z(\cl)$
normalization factors, With a joint distribution, there is no need to
compute the $\Z$ normalization factor and $\argmax_\cl \Pgibbs(\x,\cl) =
\argmax_\cl \lbd \cdot \fvec(\x, \cl)$.}
$
\argmax_\cl \left( \lbd \cdot \fvec(\x, \cl) - \ln \Z(\cl) + \ln \Pemp(\cl)
\right)
$.

\subsection{Estimating conditional distributions}

In this section, we propose a novel extension to the \slmax\ algorithm to estimate the parameters of the
maxent model for the conditional distribution $\Pest(\cl|\x)$.
In the literature~\cite{nigam,och}, conditional maxent  is typically
the only distribution considered 
for classification, as it is expected to be the most discriminant.
However, its optimization turns out to be more complex, so we present it last.

In the case conditional distributions, the main challenge is that, for each training sample $i$, we want to
estimate one separate distribution over the classes $\Pemp_i$. At the same
time, the constraints apply to the entire training set and tie up these
distributions. If we trained each distribution separately for each sample $i$,
constraints would be $\forall \clp,j: \abso{ \Pest_i[\feat] - \Pemp_i[\feat]}
\leq \beta_{\clp,i,j}$. This would result in $\Nx \times \Nfeat \times \Nclass
$ learnable parameters, with obvious overfitting. On the other hand, summing these constraints
over the examples produces $\abso{\frac{1}{\Nx}\sumi \Pest_i[\feat] -
\Pemp[\feat]} \leq \betaj$. This formulation was used before
\cite{nigam}, but we added regularization and a different sequential-update
algorithm.

The two optimization problems are:
\optp{\ppost}
{ \maxl_{\Pest_1, \dots, \Pest_{\Nx}} \sumi \entr{\Pest_i}}
{
\forall i : \sumcl \Pest_i(\cl)=1\\
\forall \clp,j: \abso{\frac{1}{\Nx}\sumi \Pest_i[\feat] - \Pemp[\feat]} \leq \betaj
}
\optq{\qpost}
{\minl_{\lbd} \loss(\lbd)}
{
\loss(\lbd)= -\frac{1}{\Nx}\sumi \Pemp_i[\ln{\Pgibbs}]+\sumdj \betaj \abso{\lambda_{\clp,j}}\\
\Pgibbs(\x,\cl)= \frac{1}{\Z(\x)}\Ogibbs{\x, \cl}\\
\Z(\x)=\sum_{\cl} \Ogibbs{\x, \cl}
}

The likelihood can be expanded as
\begin{multline}
\loss(\lbd)= -\sumdj \lambda_{\clp,j} \Pemp[\feat] + \frac{1}{\Nx}\sumi \ln
\Z(\x_i)\\
 +\sumdj \betaj \abso{\lambda_{\clp,j}}.
\end{multline}
The novelty in problem  $\qpost$ involves having one normalization constant per
example. In the development of the  likelihood, a single logarithm $\ln \Z$ is
replaced by the sum $\frac{1}{\Nx} \sumi \ln \Z(\x_i)$.
To bound $\loss(\lbd')-\loss(\lbd)$, the most difficult step is to bound:
\begin{align}
\Delta &=    \frac{1}{\Nx}\sumi \ln \frac{Z_{\lbd'}(\x_i)}{\Z(\x_i)}\\
& =\frac{1}{\Nx}\sumi \ln  \left( \sum_{\cl\neq \clp} \Pgibbs(\x_i,\cl) + \Pgibbs(\x_i,\clp)
e^{\delta\feat(\x_i,\clp)}\right)\\
&\leq   \frac{1}{\Nx} \sumi \ln\left(1+(e^\delta-1)
\Pgibbs(\x_i,\clp)\feat(\x_i,\clp)\right)\label{eq:binary}\\
&\leq
\ln\left(1+(e^\delta-1) \Pgibbs'[\feat] \right)\label{eq:jensen}
\end{align}
where $
\Pgibbs'[\feat] =\frac{1}{\Nx} \sumi \Pgibbs(\x_i,\clp)\feat(\x_i,\clp)
$.

\equat{binary} uses $$
e^{\delta\feat(\x_i,\clp)} \leq 1+
(e^{\delta}-1)\feat(\x_i,\clp)
$$
 for $\feat(\x_i,\clp) \in [0,1]$ with
equality if $\feat(\x_i,\clp) \in \{0,1\}$.
\equat{jensen} relies on the convexity of the log function to apply Jensen's
inequality. 

We have established here a new bound on the decrease in the
loss for L1-regularized {\em conditional} maxent models:
\begin{multline}
\label{eq:nbound}
\loss(\lbd')-\loss(\lbd) \leq -\delta \Pemp[\feat] + \ln\left(1+(e^\delta-1)
\Pgibbs'[\feat]\right)\\
 + \betaj \left(\abso{\lambda_{\clp,j}+\delta} - \abso{\lambda_{\clp,j}}\right).
\end{multline}
Because of its similarity to the standard \slmax\ bound (\equat{nbound}), it allows a simple
generalization of the \slmax\ algorithm to conditional distributions by
replacing the $\Pgibbs[\feat]$ with $\Pgibbs'[\feat]$.
Our experiments in the case of binary features show that the bound
given in \equat{bound} is very tight. It  can be used to obtain, in a
closed form manner, a value for $\delta$ that is close to the optimum.

In the case of conditional maxent, it is instructive to
compare this algorithm to Improved Iterative Scaling(IIS)~\cite{berger}, which
also updates the parameters to maximize a bound on the decrease in the
loss.  First, the bounds are significantly different. The \slmax\ bound is
tighter because it requires only a single $\lambda_j$ to be modified at a
time.  Second, while both approaches support closed form solutions under
specific conditions, these conditions are very different: the features must be
binary in the case of \slmax, and they must add up to a constant value in the
case of IIS. Finally, to our knowledge, there is no simple modification of IIS to
handle L1-regularization.

\ignore{
\subsection{Choosing the regularization constant}

In the original SL1-Max algorithm, \emcite{maxent-colt} suggest using $\beta_j = \beta
\frac{\Semp[f_j]}{\sqrt{m}}$ where $\Semp[f_j]$ is the standard deviation of
$f_j$ under $\Pemp$. This heuristic is easy to adapt to the three
distributions we have described so far.

As a sanity check, we verify that the ratio between the betas and the feature
empirical means remain the same.  If we assume binary features, it is easy to verify that
$\Semp[f_j]=\sqrt{\Pemp[f_j] (1-\Pemp[f_j])}$

With the joint distribution,
we have $$
\betaj(joint)=\beta \frac{\Semp[\feat]}{\sqrt{\Nx}}=
 \beta \sqrt{\frac{\Pemp[\feat] (1-\Pemp[\feat])}{\Nx}}.
$$
With the class-conditional distribution, the size of the sample set for class
$\cl$ is only $\Pemp(\cl) \Nx$. Over this reduced sample set, we have larger
class-conditional feature empirical means
$\Pemp_\cl[\feat]=\frac{\Pemp[\feat]}{\Pemp(\cl)}$, and expect $\beta$ to
scale with the same $\frac{1}{\Pemp(\cl)}$ factor.
It is easy to verify that
$$
\betaj(cond)=\beta\frac{\Semp_\cl[\feat]}{\sqrt{\Pemp(\cl) \Nx}}= 
\frac{\betaj(joint)}{\Pemp(\cl)} \sqrt{\frac{1-\Pemp[\feat]}{1-\Pemp_{\cl}[\feat]}}
$$
With sparse data, where $\Pemp_{\cl}[\feat] \ll 1$ and $\Pemp[\feat] \ll 1$,
we have $\betaj(cond) \approx \frac{\betaj(joint)}{\Pemp(\cl)}$, which shows
that the heuristic for choosing $\beta$ is consistent.
}

\section{Multi-label categorization}
\label{multi}

The fundamental modeling assumption we have made so far implies that each example $i$
only carries a single label $\cl_i$. However, in many classification problems,
a given input can correspond to multiple labels (multi-label).

For simplicity, we assume that there is  no form of ranking or
preference~\cite{aiolli} among the labels.
Our sample space covers input-code pairs $(\x, \cvec) \in \xset \times
\{0,1\}^{\Nclass}$, where $\cvec$ is a binary output code.
\ignore{
 For a given input $\x$, there can be different types of
assumptions:
\bd
\item[Label uncertainty] For a given input $\x$, determine the most likely class $\cl^*=\argmax
  _\cl \Ptrue(\cvec[\cl]=1|\x) $. In this
  case, we may also assume that the multiple labeling for a training example
  represents uncertainty about the class. 
\item[Label multiplicity] For a given input $\x$, determine the binary vector $\cvec$. During
  training, minimize the Hamming error between the output vector 
  $\cvec$ and the target vector.
\ed
We first propose two straightforward extensions to the models we have
  described so far, and show their limitations. Then, a formal solution that encodes the target classes as a binary
vector is considered.
}

Class-conditional distributions represent the easiest way to deal with
multiple labels. We only focus on the estimation of $\Pemp_\cl$ and the fact that there are
multiple labels that can be ignored. However, the final classification
decision will require a multiplication by $\Pemp(\cl)$ that is not defined as a probability
distribution because $\sum_{\cl} \Pemp(\cl)>1$.

This section reviews two other techniques to handle multiple labels that only
require minimum modifications of the algorithms proposed so far and show
their limitations. 
\ignore{More complex techniques inspired by the ranking
literature~\cite{aiolli} are possible, and currently being investigated.}

\subsection{Duplicating training examples}
\label{cond_mclass}
Assume the training data is a set of input-code pairs $(\x_1, \cvec_1), \dots,
(\x_{\Nx}, \cvec_{\Nx})$.
Our goal is to project this training set in the smaller input-label sample space $\{\x_1, \dots,
  \x_\Nx\} \times \{1, \dots, \Nclass\}$. For each training sample $(\x_i, \cvec_i)$ in the
  input-code space , we build $K_i$ samples in the input-label  space with
  $K_i=|\{1\leq k \leq \Nclass: \cvec_i[k]=1 \}|$ The conditional probability function is:
\beq
\Pemp(\cvec[k]=1|\x_i)= \left\{
\begin{array}{ll}
\frac{1}{K_i} & {\rm if\ } \cvec_i[k]=1 \\
0 & {\rm otherwise}
\end{array}
\right.
\eeq

The problem with this approach is that the empirical distribution $\Pemp$ is
reweighted to favor examples with multiple labels with $\Pemp(\x_i) =
\frac{K_i}{\sum_i K_i}$ (we assume here that $\forall i,j : \x_i \neq \x_j$).

\ignore{If we reweight each example derived from $\x_i$ with $\frac{1}{K_i}$, then
$\Pemp(\x_i) = \frac{1}{\Nx}$.  In this case, if $K_i=0$, example $i$ will be
ignored.}

\subsection{Using a non-class model}
\label{nonclass}

With an output code that represents $\Nclass$ binary decisions, a trivial
solution is to build  $\Nclass$ binary classifiers. This is typically what
$\Nclass$ \vsother\ classifiers do, and, in the case of maxent, one may
think that the $\Nclass$ class-conditional classifiers described in
\sect{classcond} represent such a solution. 

However, the statistical reality is more complex; an independence
assumption between each binary output is necessary:
\beq
\label{eq:ind}
\Ptrue(\cvec|\x) = \Ptrue(\celem^{1}, \dots, \celem^{\Nclass}|\x) = \prod_\cl \Ptrue(\celem^{\cl}|\x)
\eeq
Under this assumption, one can estimate each $\Ptrue(\celem^{\cl}|\x)$
independently. For each class $\cl$, we introduce the  distributions
$\Pest^{\cl}_i(y)= \Pest^{\cl}(y|\x_i)=\Pest(\celem^{\cl}=y|\x_i)$  where $y \in \{0,1\}$ is the the
index for a secondary classification problem between examples belonging to
class $\cl$ and the other examples (which are said to be part of $\cl$ {\em non-class}).
The independence assumption can be rewritten as 
$\Pest_i(\cvec)= \prod_\cl \Pest^{\cl}_i(\celem^{\cl})$ so that
the overall entropy can be decomposed into one entropy per class:
\beq
\entr{\Pest_i}
= -\sum_{\cvec \in \{0,1\}^{\Nclass}} \Pest_i(\cvec) \ln \Pest_i(\cvec) 
= \sum_{\cl} \entr{\Pest^{\cl}_i}
\eeq
As the entropy of each of the $\Pest^c$ distribution entropies can be
maximized separately, we have $\Nclass$ binary maxent models to
estimate. Conditional maxent has previously been applied to binary
``question answering'' problems~\cite{nonclass}.

The framework defined by problems $\ppost$ and $\qpost$ to produce the set of
parameters $\lbd$ can be applied here to the distribution $\Pest^{\cl}$.
The transformation of notation  is described in the following table:\\
\begin{tabular}{c||c|c}
\hline
Distribution & $\Pest$& $\Pest^{\cl}$\\
\hline
Problem & $\qpost$ & $\qpostp$\\
\hline
Parameters & $\lbd$ & $\lbd^{\cl}$\\
\hline
Features & $\fvec(\x,\cl)$ & $\fvec^{\cl}(\x,y)$\\
with & $\cl \in \{1,\dots, \Nclass\}$ & $y \in \{0,1\}$\\
\hline
\end{tabular}

The exponential distribution that solves
the dual problem $\qpostp$ takes the form:
$$
q_{\lbd^\cl}(\x,y)= \frac{1}{Z_{\lbd^{\cl}}(\x)} e^{\lbd^{\cl} \cdot
  \fvec^{\cl}(\x,y)}
$$
The simplifying assumption of \equat{simpli} becomes here:
\beq
f_{y',j}^\cl(\x,y)= \left\{
\begin{array}{ll}
v_j(\x) & {\rm if\ } y= y' \\
0 & {\rm otherwise}
\end{array}
\right.
\eeq

Thus $\lbd^{\cl} \cdot \fvec^{\cl}(\x,y)= \lbd_y^{\cl} \cdot \mathbf{v}(\x)$ and
the probability of observing class $\cl$ becomes
\beq
\label{eq:logit}
q_{\lbd^\cl}(\x,y)=  \sigma \left( (\lbd_1^{\cl} - \lbd_0^{\cl})
\cdot \mathbf{v}(\x) \right)
\eeq
where $\sigma(x)=\frac{1}{1+e^{-x}}$ is the sigmoid function. Note that for each
feature $v_j(\x)$, classifier $c$ has two parameters: $\lambda^c_{1,j}$
used in the positive model and $\lambda^c_{0,j}$
used in the negative model.
Given a test input $\x$, this  approach can be used either to produce the code
vector $\cvec$ such that $q^{\cl}_{\lambda}(\x,\celem^{\cl})> 0.5$ or the top
class $\argmax_\cl q^{\cl}_{\lambda}(\x,1)$.

\equat{logit} suggests that, in the binary case, conditional
maxent amounts to logistic regression. The use of L1-regularization
in logistic regression was recently analyzed \cite{ng}. Another technique
to optimize the logistic loss that relies on an implicit L1 regularization is
AdaBoost with logistic loss~\cite{collins00}, which also uses a sequential
update procedure similar to \slmax.

While the independence assumption is not as straightforward, we can also transpose
problem  $\pjoint$ to the distributions  $\Pest^{\cl}(\x,y)=\Pest(\x_i,\celem^{\cl}=y)$ 
and obtain the convex dual $\qjointp$.

\begin{table}[t]
\begin{center}
\begin{tabular}{|c|c|c|c|c|}
\hline
Assume & {\small Implement} & Maxent & Optim. & Z?\\
Classes & {\small Classifiers} & Distrib. & problem & \\
\hline
\hline
 & & {Joint} & $\qjoint$ & No \\
\cline{3-5}
{\bf Exclu-} & {\bf Tied} & {Conditional} &  $\qpost$ & No\\
\cline{2-5}
{\bf sive} & & {ClassCond} & $\qclass$ & Yes\\
\cline{1-1}\cline{3-5}
& & {Joint} & $\qjointp$ & No\\
\cline{3-5}
{\bf Inde-} & {\bf Sepa-} & {Conditional} &  $\qpostp$ & No\\
\cline{3-5}
{\bf pendent}  & {\bf rate}&  \multicolumn{2}{|c|}{\boost} & No\\
\hline
\end{tabular}
\end{center}
\caption{Impact of the model on the implementation. The last column indicates that
  the computation of the normalization Z is required to perform classification.}
\label{tab:implement}
\end{table}
 
\section{Implementation: it's all about normalization}
\label{implement}
This section shows that from an implementation viewpoint, normalization is the
main differentiator in the algorithms described in this paper.

We have already noted than SL1-Max is strikingly similar to AdaBoost,
especially when AdaBoost is described within the Bregman distance
framework~\cite{collins00}. As a matter of fact, unregularized conditional
maxent was shown~\cite{lebanon} to be equivalent to 
AdaBoost with the additional constraint of $\sumcl \Pest_i(\cl)=1$. 
 
Our implementation of SL1-Max capitalized on an earlier implementation
of AdaBoost to include a normalization constant. The joint model
requires normalization over all the classes and examples. The
class-conditional model requires, for a given class, normalization over all
the examples. The conditional model requires, for a given example,
normalization over all the classes.

\ignore{
The correspondence between  AdaBoost with logistic loss and class-independent
maxent is more complex, as AdaBoost does not require a separate {\em
  non-class} model.

We will see however in our experimental section that while AdaBoost leads to a
sparser solution, its learning convergence is significantly slower.
}
Our implementation of  multiclass \slmax, which derives directly from
Sections~\ref{general} and \ref{nonclass}, can be interpreted as a
combination of \vsother\ classifiers. We have not explored output codes or
hierarchical structures, though multiclass \slmax\ offers a promising framework
to explore these approaches, where the class-independent hypotheses would be
much closer to reality. The most important difference between the various
modeling hypotheses is whether the classifiers can be trained separately,
or are tied by shared normalization constants. Training classifiers
separately can be done on parallel computers or sequentially; in either case,
the implementation is more memory efficient in a way which is
critical when the  number of classes is very large. 

There is some merit in training the classifiers together: one can minimize a
unique target function and monitor the training process in a variety of
ways. The most common stopping criterion is when the classification error
minimum is reached on validation data. When training classifiers separately,
the absence of a single stopping criterion makes the process much harder to
monitor.
\tbl{implement} summarizes the merit of each modeling assumption from an
implementation viewpoint.

An implementation of \slmax\ that is optimized using {\em partial pricing}
strategies~\cite{pricing} is provided in the (blanked out) software
package. When classifiers can be trained separately, an \slmax\ binary
classifier is just another \vsother\ classifier that can be used instead of an
AdaBoost or a SVM classifier. On a given classification learning task,
choosing between \slmax, AdaBoost and SVM can be done with a single switch,
or by automatically using cross-validation data. Systematic experimental
comparisons between the three approaches for large scale natural language
understanding tasks~\cite{spcom} indicate that \slmax\ is the fastest
approach on datasets larger than 100,000 examples, with state-of-the-art
accuracy\footnote{They are only outperformed by SVMs with polynomial kernels,
which are not a computationally practical because of the large number of
support vectors}.

It would be informative to compare \slmax\ to algorithms considered as the
state-of-the art for the estimation of parameters in conditional entropy
models. They include Iterative Scaling algorithms, such as IIS~\cite{berger}
and Fast Iterative Scaling(FIS)~\cite{jin03faster}, and gradient
algorithms~\cite{malouf}. This study, which would be of considerable
interest, is beyond the scope of this paper.

The two key factors that contribute to the remarkable learning speed of
\slmax\ have not been, to our knowledge, applied to most algorithms in the
iterative scaling family. First, \slmax\ is based on a pricing strategy:
modify the single parameter which causes the greatest decrease in the
objective function. The addition of {\em partial} pricing can make the search
for the parameter considerably faster.  Second, the L1 regularization adds
some slack in the constraints and makes them easier to satisfy early in the
optimization process.

Results on the WebKB text classification task show
that \slmax\ takes less than 10 seconds to learn 3150 examples with 25,000
features, which compares favorably to more than 100 seconds when using FIS or
IIS on a reduced set of 300 features~\cite{jin03faster}. (we assume comparable
  Pentium CPUs with a 2GHz clock).
\ignore{(our CPU is 2.4GhZ Pentium).}

\begin{table}[t]
\begin{center}
\begin{tabular}[t]{|c||c|c|c|}
\hline
 & Reuters & WebKB & SuperTags \\
\hline
Multi-label? & Yes & No & No\\
Train size & 9603 & 3150 & 950028 \\
Test size & 3299 & 4199 & 46451 \\
Num. of labels & 90 & 4  & 4726 \\
Num. of features& 22758 & 25229 & 95516 \\
Features/sample & 126.7 & 129.1 &  18.8 \\
\hline
\end{tabular}
\end{center}
\caption{Key characteristics of the three datasets used in the
experiments. The last line gives the
average number of non-zero features per training vector.}
\label{tab:data}
\end{table}

\section{Experiments and Discussions}
\label{expe}

The first two datasets are small enough to allow us to run the methods $\qjoint$
and $\qpost$, which are compared to $\qclass$, $\qjointp$, and $\qpostp$. The
Reuters-21758\footnote{\tt \tiny http://www.daviddlewis.com/resources/testcollections/reuters21578} dataset contains stories collected from Reuters newswire in
1987. We used the ModApte split between 9603 train stories and 3299 test
stories. This is a multi-label problem, where the number
of labels per story ranges from 0 to 15. The WebKB\footnote{\tt http://www-2.cs.cmu.edu/~webkb} dataset
contains web pages gathered from university computer science departments. We
selected the same categories as~\cite{nigam}: {\em student}, {\em faculty},
{\em courses} and {\em projects}. The 4199 samples are split between
training and testing using a 4-fold cross-validation.

The third dataset, which demonstrates the scaling ability of \slmax, is much
larger; and only the methods $\qclass$, $\qjointp$, and $\qpostp$ can be
applied. It consists of a set of SuperTags. SuperTags are extensions of
part-of-speech tags that encode morpho-syntactic
constraints~\cite{srinijoshi99}) and are derived from the phrase-structure
annotated Penn TreeBank. The characteristics of the three datasets are
summarized in \tbl{data}.

\begin{table}[t]
\begin{center}
\begin{tabular}[t]{|c||c|c|c|}
\hline
 & Reuters & WebKB & SuperTags \\
\hline
AdaBoost & 14.9/86.7   & 7.10 &  12.0 \\
Linear SVM & 15.3/86.6   & 7.57 &   \\
\hline
$\qjoint$ & 16.9/84.0 & 7.95 &  \\
$\qclass$ & 17.4/83.9 & 8.19 &  11.2 \\
$\qpost$ & 16.6/79.6 & 7.76 &  \\
\hline
$\qjointp$ & 15.0/86.5 & 7.50 &  11.9 \\
$\qpostp$ & 14.1/86.4 & 7.45 &  11.1\\
\hline
\end{tabular}
\caption{Error rates on the 3 datasets. For Reuters, which is  multi-label, the first number is the top-class error
rate (an example is considered an
error if the highest scoring class given by the classifier is not part of the
target labels) and the second number is the micro-averaged optimal F-measure.}
\end{center}
\label{tab:res}
\end{table}

Table 3 compares the five different multiclass \slmax\ models considered in
Sections~\ref{general} and \ref{nonclass}. The \slmax\ regularization
parameter is set to $\beta=0.5$.  AdaBoost with logistic loss and linear SVMs
are provided as a baseline (note that our implementation of AdaBoost can be
considered as a class-independent model with separate classifiers). The best
error rate we obtain on WebKB (7.1\%) and the best F-Measure we obtain on
Reuters (86.7\%) compare favorably to the
literature~\cite{nigam,joachims98}. The training speed of Adaboost and
\slmax\ are very similar and can be optimized using the same techniques. They
are not reported here as a detailed comparison is reported
elsewhere~\cite{pricing}.  \ignore{ In the
case of the SuperTags set, the learning convergence of AdaBoost was extremely
slow (more than one order of magnitude slower than \slmax) and its inferior
performance can be explained by the fact that learning did not fully converge
(this would have taken weeks).  }

{\bf How good are class-conditional models?}  The most computationally
efficient model is the {\em class-conditional} model $\qclass$. Table 4, which
compares the computational efficiency of the $\qclass$, $\qjointp$, and
$\qpostp$ models, shows that it has the smallest training time and the smallest
number of parameters.  However, its error rate on smaller dataset is higher due to the estimation
of the $Z$ normalization constant.

{\bf Class-exclusive vs. independent assumptions:} The ``Reuters''
column of Table 3 indicates that making a class-exclusive assumption when it
is not justified (e.g. the Reuters data is multi-label) leads to a significant
loss in performance. By contrast, the class-independent assumption is never
true, but a combination of binary maxent classifiers, which relies on this
assumption, consistently improves performance. It also greatly improves
training speed by allowing parallelization and a small memory footprint. A
combination of binary classifier yields excellent {\em classification accuracy}
regardless of the size of the problem. However, comparisons on a ``Question
Answering'' problem~\cite{nonclass} suggest that they may not perform well for
{\em ranking} tasks. Future work on multi-label tasks will also assess the ranking performance with
specific error measures~\cite{aiolli}.

{\bf Pros and cons of conditional models:}
For pure classification, the fully discriminant {\em conditional} models
($\qpost$ and $\qpostp$) yield the best results. This may justify the
exclusive use of conditional models in all previous studies of multiclass
maxent. Table 4 shows another reason to prefer conditional models:
they are {\em sparser} and require fewer model parameters (since
they only focus on performing class discrimination). 
Conditional models have two major drawbacks. First, as their outputs are
normalized separately for each example, they tend to be poor confidence 
estimators. Metrics based on the comparison of the output to a varying
threshold tend to fare poorly, For instance, in Table 3, the conditional
model $\qpost$ yield the lowest F-measure for Reuters (79.6).
Second, as shown in Table 4, they can require more
training time.

\begin{table}[t]
\begin{center}
\begin{tabular}[t]{|c||c|c||c|c|}
\hline
 & \multicolumn{2}{|c||}{\# parameters} & \multicolumn{2}{|c|}{Train time} \\
& \multicolumn{2}{|c||}{(thousands)} & \multicolumn{2}{|c|}{(hours)} \\
\hline
$\beta$  & 0.5 &  0.9 & 0.5 &  0.9 \\
\hline
$\qjointp$ & 53 &  35 & 15.65 & 16.09 \\
$\qclass$ & 53 &  34 & 14.92 &  7.58 \\
$\qpostp$ & 41 &  24 & 46.65 & 46.40 \\
\hline
\end{tabular}
\caption{Number of non-zero model parameters and training time for the SuperTags set as a function of
  the regularizer $\beta$ and the optimization method. On this large set,
  $\beta$ has little impact on accuracy, and mostly affects speed and sparsity.}
\end{center}
\label{tab:feat}
\end{table}

\section{Conclusions} 
We have shown that a sequential maxent algorithm (SL1-Max) can be applied to
many classification problems with performances which are comparable to
Adaboost and SVMs. An important (and apparently under-appreciated) advantage of maxent for classification problems
appears to be its remarkable flexibility in terms of modeling
assumptions. In future work, this flexibility will be used to optimize maxent for problems
where ranking or rejection performances are critical, and to which traditional
classification methods are problematic to adapt.

\bibliographystyle{plain}
\bibliography{all}
\end{document}